\documentclass[letterpaper, 10pt, conference]{ieeeconf}
\IEEEoverridecommandlockouts
\usepackage{graphicx,epstopdf,multirow,multicol,subfigure,booktabs,pgfplots}

\let\proof\relax
\let\endproof\relax
\usepackage{amsmath,bm}
\usepackage{amssymb,amsthm}
\usepackage{times}
\usepackage[libertine]{newtxmath}
\usepackage[colorlinks]{hyperref}
\usepackage[T1]{fontenc}
\usepackage{booktabs}
\usepackage{cite}
\usepackage{soul}
\allowdisplaybreaks
\pgfplotsset{compat=1.13}

\newtheorem{proposition}{Proposition}

\title{\LARGE \bf
Path Planning for Multiple Heterogeneous Unmanned Vehicles with Uncertain Service Times }

\author{Kaarthik Sundar$^{\dagger}$\thanks{$^{\dagger}$ Center for Non-Linear Studies, Los Alamos National Laboratory, Los Alamos, NM. \texttt{kaarthik01sundar@gmail.com}},\;
Saravanan Venkatachalam$^{*}$\thanks{$^{*}$ Assistant Professor, Dept. of Industrial Engg., Wayne State University, Detroit, MI. \texttt{}},\;
Satyanarayana G. Manyam$^{\ddagger}$\thanks{$^{\ddagger}$ National Research Council Fellow, Air Force Research Laboratory, WPAFB, OH.}} \;

\renewcommand{\baselinestretch}{1.2}

\begin{document}

\maketitle

\begin{abstract}
This article presents a framework and develops a formulation to solve a path planning problem for multiple heterogeneous Unmanned Vehicles (UVs) with uncertain service times for each vehicle--target pair. The vehicles incur a penalty proportional to the duration of their total service time in excess of a preset constant. The vehicles differ in their motion constraints and are located at distinct depots at the start of the mission. The vehicles may also be equipped with disparate sensors. The objective is to find a tour for each vehicle that starts and ends at its respective depot such that every target is visited and serviced by some vehicle while minimizing the sum of the total travel distance and the expected penalty incurred by all the vehicles. We formulate the problem as a two-stage stochastic program with recourse, present the theoretical properties of the formulation and advantages of using such a formulation, as opposed to a deterministic expected value formulation, to solve the problem. Extensive numerical simulations also corroborate the effectiveness of the proposed approach.
\end{abstract}

\section{Introduction \label{sec:intro}}
Advances in sensing, robotics, and wireless networks have enabled the use of teams of Unmanned Vehicles (UVs) for environmental sensing applications including crop monitoring \cite{Shanahan2001}, ocean bathymetry \cite{Ferreira2009}, forest fire monitoring \cite{Casbeer2005}, ecosystem management \cite{Corrigan2007}, civil security applications such as border surveillance, and military applications such as reconnaissance and data collection missions. These applications frequently require vehicles to collect data such as visible/infra-red/thermal images, videos of specified target sites, and environmental data such as temperature, moisture, humidity using on-board sensors, and deliver them to a base station. Typically, in these applications, path planning for the UVs has to be performed a priori before the start of the mission in the presence of uncertainty in travel times, service times, communication delays etc. This paper considers a stochastic generalization of a fundamental path planning problem involving a fleet of multiple heterogeneous vehicles. We classify the \emph{heterogeneity} of these vehicles into two categories: \emph{structural} and \emph{functional} heterogeneity. Vehicles are said to be structurally heterogeneous if they differ in design and dynamics. This can lead to differences in maximum speed at which they can travel, payload capacity, fuel consumption, etc \cite{Sundar2012, Sundar2014, Levy2014, Sundar2016c, Sundar2016d}. This is a rational assumption as some structural differences are always present between any pair of vehicles. A collection of vehicles is said to be functionally heterogeneous if not all vehicles may be able to service a target. The vehicles may be equipped with disparate sensors due to payload restrictions, and it leads to functional heterogeneity. In this case, we partition the set of targets into disjoint subsets: (a) targets to be visited by specific vehicles, and (b) common targets that can be visited by any of the vehicles.

%\subsection{Problem Definition} \label{subsec:prob_defn}
The problem we consider in this article is motivated by a mission scenario, where a collection of UVs have to visit a set of target locations and collect data. The UVs have to establish a communication link with sensors located at the target locations and transfer the sensor data. The time required to establish the communication link and complete the data transfer is uncertain, and is modeled as a random variable with a known distribution. To address any such routing problem with uncertain service times, we pose the following heterogeneous, multiple depot, multiple unmanned vehicle path planning problem with random service times (HMDMVPP-RST). We are given (i) a set of targets, which are locations of ground sensors collecting data, (ii) a fleet of heterogeneous vehicles located at distinct depots, a heading angle for visiting each target and each depot, (iii) an uncertain service time for every vehicle--target pair, and (iv) a preset limit for every combination of  vehicle--target that could be indicative of the maximum time the vehicle takes to service the target. The objective is to find a tour for each vehicle that starts and ends at its depot such that each target is visited by at least one vehicle, the vehicle--target constraints are satisfied, the paths satisfy the motion constraints of the respective vehicles, and sum of the travel cost by all the vehicles and the total expected penalty is minimized. The penalty incurred for a vehicle is proportional to the duration of its total service time in excess of a preset limit; this limit may be considered as a budget on the total service time for each vehicle. Typically, the time taken for a vehicle to establish a reliable communication link with the ground sensor at a given target and collecting the data from it varies based on a plenty of natural factors like wind, obstacles etc. The vehicle is assumed to loiter at a target until a reliable communication link is established with the ground sensor and the data transfer is completed; this loitering time (service time) at a target is uncertain. 

%This uncertainty in service time at vehicle--target pair is modeled as a random variable with a known distribution. 
%(service time at a target for a given vehicle refers to the time taken for a vehicle to establish a communication link with the ground sensor at the target and collect the data)

%\subsection{Assumptions} \label{subsec:assumptions}
To formulate the HMDMVPP-RST, we make the following assumptions: each vehicle in the heterogeneous fleet of UVs is modeled as a Dubins vehicle with distinct value for its minimum turn radius and all the vehicles are assumed to travel at a constant velocity. The paths for all the vehicle are determined before knowing the realizations of the random service times. The UVs must follow their a priori paths; no path re-optimizations are permitted. The uncertainty in the service time for each vehicle--target pair is available in the form of samples or scenarios from a known distribution. In the presence of uncertainty, it is often practical to have recourse decisions. The optimal paths for all the vehicles are determined while hedging against the uncertainty before the realization of randomness in service time.

Following are the contributions of this paper: we develop a two-stage stochastic formulation for the HMDMVPP-RST which explicitly incorporates the modeling of the uncertainty in service times for each vehicle--target pair. We develop a branch-and-cut algorithm which an algorithm that can provide an optimal solution to any instance of the problem. We present the conditions under which the two-stage formulation and a deterministic expected value based formulation that uses the expected values for the uncertainty in service times are equivalent for a particular class of instances. Finally, extensive computational experiments are presented illustrating the advantages of a two-stage stochastic formulation over the deterministic expected value counterpart.

\subsection{Related Work} \label{subsec:related_work}
The deterministic variant of the HMDMVPP-RST is a generalization of the asymmetric multiple vehicle heterogeneous traveling salesmen problem \cite{Sundar2016} which is NP-hard. The authors in \cite{Sundar2016, Sundar2015} formulate and develop algorithms that can compute an optimal solution for the symmetric and asymmetric multiple vehicle heterogeneous traveling salesmen problems, respectively. Another variant of the HMDMVPP-RST where the vehicles are homogeneous and modeled as point masses is the stochastic multiple vehicle routing problem with random travel times \cite{Kenyon2003}. Authors in \cite{Laporte1992a, Li2010} consider a vehicle routing problem with uncertainties in both service times and travel times; in particular \cite{Li2010} also includes time windows during which the vehicle should visit the targets. In \cite{Laporte1992a}, the authors consider multiple homogeneous vehicles, whereas, this paper considers a heterogeneous version of the problem with motion constraints. A number of variants of the deterministic version of the HMDMVPP-RST involving one or more vehicles have been addressed in the literature and a plethora of heuristics and algorithms to obtain an optimal solution to the respective problems have been developed. An interested reader is referred to an excellent survey article \cite{Laporte1992} on all the deterministic variants. 

The stochastic variants of the HMDMVPP-RST are also plenty in number; interested readers are referred to a very recent literature survey on the models and techniques used to formulate and solve such variants \cite{Oyola2016}. To the best of our knowledge, none of the stochastic variants in the literature consider a heterogeneous fleet of UVs differing in their motion constraints; this is the first work in the literature to address such a stochastic variant. But there are many homogeneous variants that consider more complicated recourse actions which involve route re-planning. The focus of this paper is to describe a framework to formulate and solve problems concerning path planning for multiple \emph{heterogeneous} UVs in the presence of uncertainty in service times and illustrate the advantages of such a framework. Furthermore, developing algorithms to solve a such two-stage stochastic formulation with recourse is considered as a stand-alone area in the optimization literature \cite{Benders1962,Wollmer1980,Gendreau1996,Toth2014,Beier2015,Venkatachalam2014}. We also remark that this framework presented here for the HMDMVPP-RST is not new and has been extensively studied in the optimization literature for problems concerning decision making under uncertainty \cite{Birge2011} in supply chain management, revenue management, vehicle routing etc. 

The rest of the paper is organized as follows: the Sec. \ref{sec:notations} presents detailed notations that are used to formulate the HMDMVPP-RST, followed by the two-stage stochastic formulation in Sec. \ref{sec:formulation}. Sec. \ref{sec:algorithm} develops a branch-and-cut algorithm to solve the formulation in Sec. \ref{sec:formulation} to optimality and finally, Sec. \ref{sec:results} presents extensive computational results followed by conclusions and possible extensions in Sec. \ref{sec:conclusion}.

\section{Notations and uncertainty model} \label{sec:notations}
This section introduces the required notation to develop the two-stage stochastic formulation for the HMDMVPP-RST. To that end, let $T$ denote the set of targets and $D = \{d_1, \dots, d_n\}$ the set of depots; we have a heterogeneous fleet of $n$ vehicles initially stationed at distinct depots. We shall refer to $V:= D\cup T$ as the set of vertices. Associated with each vertex $i \in V$ is an orientation angle $\theta_i$. $\theta_i$ is the angle at which any vehicle has to arrive and depart from the vertex. Furthermore, we also assume that there are vehicle--target constraints where each vehicle $k$ is required to visit a subset of targets $R_k \subseteq T$ with $\cap_i R_i = \emptyset$; $R_k$ denotes the functional heterogeneous targets that the vehicle $k$ has to visit. The sets $R_1, \dots, R_n$ are assumed to be known a priori, and targets in the set $T\setminus (\cup_i R_i)$ are referred as common targets, and they can be visited by any vehicle. 

Each vehicle in the heterogeneous fleet of UVs is modeled as Dubins vehicle with a distinct value for its minimum turn radius. The kinematic constraints of the vehicle stationed at depot $d_k$ is given by: $v_x = v\cos \theta, v_y = v \sin \theta, \dot{\theta} \leqslant u_k$, where $v_x$ and $v_y$ are the $x$ and $y$ components of the velocities, respectively, and $\dot{\theta}$ and $u_k$ are the angular velocity and the maximum yaw-rate of the vehicle. $u_k$ is different among vehicles and $\lvert v\rvert$ is assumed to be the same for every vehicle. Since $\lvert v\rvert$ is the same for all of the vehicles, the $u_k$'s can be mapped bijectively to the vehicles' minimum turn radius values. 

Given these vehicles, the problem is formulated on a directed graph $G = (V, E)$ where $E$ is the set of directed edges joining any two vertices in $V$. We assume that the graph $G$ does not have any self-loops. For each edge $(i, j) \in E$ and each vehicle $k$, let $c_{ij}^k$ be the length of the minimum distance path for the vehicle $k$ to traverse the edge from target $i$ to target $j$ with angles of departure and arrival $\delta_i$ and $\delta_j$, respectively. This length can be computed using the well-known result of Dubins \cite{Dubins1957}. 

Also, associated with target--vehicle pair $[i,k]$ is a stochastic service time that denotes the time to establish a communication link for the vehicle $k$ with target $i$ when the vehicle visits the target $i$. Let $\bm{\tilde{\tau}} = (\tilde{\tau}_{ik})$ for every $i \in T$ and vehicle $k$ denote the non-negative random vector of service times. We use %$\Xi$ and 
$\bm{\tau}$ to denote a %the support of $\bm{\tilde{\tau}}$ and a 
realization of the random vector $\bm{\tilde\tau}$. The number of possible realizations of the service time random variable vector $\bm{\tau}$ (support of $\bm \tau$) is assumed to be finite; this is a very reasonable assumption as this is usually the only kind of data available based on past runs of the mission. Let the realizations of $\bm{\tilde{\tau}}$ be indexed by $\omega \in \Omega$ so that they can be enumerated $\bm{\tau}^\omega$, $\omega \in \Omega$, with probability mass function $\bm p^{\omega} = \operatorname{Pr}(\bm{\tilde{\tau}} = \bm{\tau}^\omega)$, $\omega \in \Omega$. We use boldface notation here and throughout the rest of the paper to denote vectors, e.g. $\bm p^{\omega} = (p^\omega_{ik})$. Also, associated with each vehicle--target pair is a preset constant, $\bar{\tau}_{ik}$, which denotes the maximum time the vehicle $k$ can service at target $i$. Finally, we let $\gamma_k$ denote a non-negative penalty per unit time of the excess duration spent on servicing any of the targets visited by vehicle $k$.

\section{A two-stage stochastic formulation} \label{sec:formulation}
In this section, we develop a two-stage stochastic formulation with simple recourse for the HMDMVPP-RST. We now define the decision variables required for the two-stage stochastic formulation of the HMDMVPP-RST. For each vehicle $k \in \{1,\dots,n\}$, we associate with each edge $(i,j) \in E$, a binary variable $x_{ij}^k$ which takes a value $1$ if the vehicle $k$ traverses the edge $(i,j)$ and $0$ otherwise. From here on, throughout the rest of the article, we denote the set of vehicles $\{1,\dots,n\}$ using $K$. Similarly for each vehicle $k \in K$ and each target $i \in T$, $y_i^k$ denotes a binary assignment variable that takes a value $1$ if the target $i$ is visited by vehicle $k$ and $0$ otherwise. For each vehicle $k \in K$, let $z_k^{\omega}$ denote the excess duration in the total service time spent by the vehicle $k$ on all the targets it visits, for the realization $\omega \in \Omega$ of the random service times. Using the above notations and those introduced in Sec. \ref{sec:notations}, the two-stage stochastic formulation for the HMDMVPP-RST is as follows:

\subsection{Objective:} 
\begin{flalign}
    &\mathcal S(\bm x, \bm y, \bm z^\omega): \min \sum_{k \in K} \left( \sum_{(i,j)\in E}  c_{ij}^k x_{ij}^k \right) + \mathbb{E}_{\bm{\tau}} \left( \sum_{k \in K} \gamma_k z_k^{\omega} \right). &\label{eq:obj}
\end{flalign}
The objective \eqref{eq:obj} minimizes the sum of the total travel distance and the expected sum of the penalties incurred by all the vehicles for the excess duration spent on servicing all the targets. Here, $\mathbb{E}$ is the expectation operator over the random variable $\bm{\tau}$. We refer to the total travel distance and the expected penalty as the first-stage cost and the second-stage/recourse cost, respectively.

\subsection{Degree constraints:}
\begin{flalign}
    &\sum_{j \in V} x^k_{ij} = y^k_i \quad \forall i \in T, k \in K  \text{ and} & \label{eq:outdegree} \\
	&\sum_{j \in V} x^k_{ji} = y^k_i \quad \forall i \in T, k \in K. & \label{eq:indegree} 
\end{flalign}
The constraints \eqref{eq:outdegree} and \eqref{eq:indegree} ensure that the out-degree and in-degree of a target that is visited by vehicle $k \in K$ is $1$, respectively.

\subsection{Sub-tour elimination constraints:}
\begin{flalign}
    &\sum_{\substack{(i,j)\in E \\i\in S, j\notin S}}x_{ij}^k \geq y^k_i \quad \forall i\in S, S\subseteq T, k \in K. & \label{eq:sec} 
\end{flalign}
 The constraints in \eqref{eq:sec} eliminate sub-tours of any subset of targets for each vehicle; they also ensure that for each vehicle, its path remains connected. They are also referred to as \emph{connectivity} constraints.
 
 \subsection{Vehicle--target assignment constraints:}
 \begin{flalign}
     &\sum_{k \in K} y_i^k = 1 \quad \forall i\in T \text{ and} & \label{eq:assignment} \\
	&y_i^k = 1 \quad \forall k \in K, i\in R_k & \label{eq:func}
 \end{flalign}
 The constraints in \eqref{eq:assignment} ensure that each target is visited by some vehicle and the constraints \eqref{eq:func} are the vehicle--target assignments for the functional heterogeneous targets.
 
\subsection{Service time constraints:}
 \begin{flalign}
    & \sum_{i \in T} \left( \bar{\tau}_{ik} - \tau_{ik}^{\omega} \right) y_i^k + z_k^{\omega} \geqslant 0 \quad \forall k \in K, \omega \in \Omega \text{ and} &\label{eq:timing} \\
    & z_k^\omega \geqslant 0 \quad \forall k \in K, \omega \in \Omega. & \label{eq:excess}
 \end{flalign}
 The term $\left( \tau_{ik}^{\omega} - \bar{\tau}_{ik} \right) y_i^k$ in constraint \eqref{eq:timing} represents the excess service time spent by vehicle $k$ at target $i$. The constraints \eqref{eq:timing} together with \eqref{eq:excess} imply that a penalty would be incurred by the vehicle only if the total duration of service time spent by the vehicle on all the targets it visits exceeds the sum of the maximum allowed service times at all of the visited targets. 
 
 \subsection{Variable restrictions:}
 {\setlength{\abovedisplayskip}{0pt}
 \begin{flalign}
     &x^k_{ij} \in \{0,1\} \quad \forall e\in E, k\in K \text{ and }& \label{eq:x} \\
	& y_i^k \in \{0,1\} \quad \forall i\in T,k \in K. & \label{eq:y}
 \end{flalign}}
The constraints in \eqref{eq:x} and \eqref{eq:y} denote the binary restrictions on the decision variables. 

We now present some interesting properties of the two-stage stochastic formulation for a single vehicle variant of the HMDMVPP-RST.

\begin{proposition} \label{prop:onevehicle}
For a single vehicle variant of the HMDMVPP-RST, the optimal path for the UV is given by an instance of the deterministic traveling salesman problem.
\end{proposition}
\proof 
For a single vehicle problem, $K=\{1\}$ and the assignment constraints in \eqref{eq:assignment} reduce to $y_i^1 = 1$ for every target $i \in T$. Hence, the $y_i^k$ variables in the the two-stage stochastic formulation can be eliminated. Then, the service time constraints in \eqref{eq:timing} reduces to $$\sum_{i \in T} \left( \bar{\tau}_{i1} - \tau_{i1}^{\omega} \right) + z_1^{\omega} \geqslant 0 \quad \forall \omega \in \Omega.$$ The above constraint does not contain any edge decision variables $x_{ij}^k$, effectively decomposing the problem to a traveling salesman problem and an expected penalty minimization problem which is given by the following linear program:
\begin{flalign*}
& \min \mathbb{E}_{\bm{\tau}} \left( \gamma_1 z_1^{\omega} \right) \text{ subject to:} & \\
& \sum_{i \in T} \left( \bar{\tau}_{i1} - \tau_{i1}^{\omega} \right)  + z_1^{\omega} \geqslant 0 \quad \forall \omega \in \Omega \text{ and} & \\
& z_1^\omega \geqslant 0 \quad  \forall\omega \in \Omega. &
\end{flalign*}
Hence, the optimal path for the single UV can be computed independently by solving a traveling salesman problem.
\endproof

\subsection{Value of Stochastic Solution (VSS)} \label{subsec:vss}
We now present a measure to compare the significance of the solution obtained by solving the two-stage stochastic problem for the HMDMVPP-RST with respect to the deterministic version of the problem, referred to as the \emph{Expected Value Problem} (EVP) in the literature \cite{Birge2011}. Given the HMDMVPP-RST, a natural predisposition is to take expectation over the service time for each vehicle--target pair and solve the following deterministic problem to obtain paths for each vehicle.
\begin{flalign}
& \mathcal D(\bm x, \bm y, \bm z): \left( \min \sum_{k \in K} \sum_{(i,j)\in E}  c_{ij}^k x_{ij}^k \right) +  \left( \sum_{k \in K} \gamma_k z_k \right) \text{ s.t.} & \notag \\
& \text{constraints \eqref{eq:outdegree}, \eqref{eq:indegree}, \eqref{eq:sec}, \eqref{eq:assignment}, \eqref{eq:func}, \eqref{eq:x}, \eqref{eq:y},} & \notag \\
& \sum_{i \in T} \left( \bar{\tau}_{ik} - \mathbb{E}_{\bm{\tau}}(\tau_{ik}^{\omega}) \right) y_i^k + z_k \geqslant 0, \text{ and } z_k \geqslant 0 \quad \forall k \in K. &\notag
\end{flalign}
The problem $\cal D$ is analogous to a single scenario, two-stage stochastic version of the HMDMVPP-RST that utilizes the expected service time for each vehicle--target pair. We remark that in the above formulation, the variable vector $\bm z$ is not a function of the realizations, $\omega \in \Omega$. Let $(\bm x^*, \bm y^*)$ denote the vector of optimal first stage decisions for the EVP that represents the optimal paths for each vehicle with the vehicle--target assignment. The VSS is a measure of how good (or more frequently, how bad) a decision $(\bm x^*, \bm y^*)$ is in terms of the two-stage stochastic problem. To that end, let $\mathcal S^*$ denote the optimal objective value to the two-stage stochastic problem in \ref{eq:obj}, and let $\mathcal D^* = \mathcal S(\bm x^*, \bm y^*, \bm z^\omega)$; $\mathcal D^*$ is the optimal objective value of the two-stage stochastic problem obtained by fixing the first stage solution $(\bm x^*,\bm y^*)$ obtained from the EVP. Then the Value of the Stochastic Solution (VSS) is given by $$\text{VSS} = \mathcal D^* - \mathcal S^*.$$ VSS provides the cost of ignoring uncertainty in choosing a decision. For any stochastic program, it is known in the literature that $\text{VSS} \geqslant 0$ \cite{Madansky1960,Birge2011}. We also note that the Prop. \ref{prop:onevehicle} can be equivalently stated as: $\text{VSS} = 0$ for the single vehicle case of the HMDMVPP-RST.

\section{Algorithm} \label{sec:algorithm}
In this section, we briefly present the main ingredients of a branch-and-cut algorithm that is used to solve the formulation in the Sec. \ref{sec:formulation} to optimality. The two-stage stochastic formulation developed in the Sec. \ref{sec:formulation} for the HMDMVPP-RST can be provided to off-the-shelf commercial branch-and-cut solvers to obtain an optimal solution. But, the formulation contains the sub-tour elimination constraints \eqref{eq:sec} for each vehicle which enforce any feasible solution to the problem to not contain any sub-tours of the targets. The number of such constraints is exponential and it may not be computationally efficient to enumerate all these constraints and provide them to these solvers. To address this issue, we use the following approach: we relax these constraints from the formulation, and whenever the solver obtains an integer feasible solution to this relaxed problem (or a fractional solution with integrality constraints dropped), we check if any of these constraints are violated by the feasible solution, integer or fractional. If so, we add the constraint \emph{dynamically} and continue solving the original problem. This process of adding constraints to the problem sequentially has been observed to be computationally efficient for the deterministic traveling salesman problem and some of its variants \cite{Sundar2016a,Sundar2016b,Sundar2016c,Manyam2016}. The algorithms used to identify violated constraints are referred to as separation algorithms. 

We shall now detail the separation algorithm used to dynamically identify violated sub-tour elimination constraint \eqref{eq:sec} given a fractional solution. For every vehicle $k \in K$, $G^{*}_k=(V_k^{*},E_k^{*})$ denotes the \emph{support graph} associated with a given fractional solution $(\bm{x}^{*},\bm{y}^{*})$, \emph{i.e.,} $V_k^{*}:=\{i\in T:y^{k*}_{i}>0\}\cup\{d_k\}$ and $E_k^{*}:=\{(i,j)\in E:x^{k*}_{ij}>0\}$. Here, $\bm{x}$ and $\bm{y}$ are the vectors of the decision variable values in HMDMVPP-RST. To check if any of the sub-tour elimination constraints in \eqref{eq:sec} are violated given an integer (fractional) solution, we first examine the strongly connected components in $G_k^*$. Each strongly connected component $C$ that does not contain the depot $d_k$ generates a violated sub-tour elimination constraint for $S=C$ and for each $i\in S$. We now define $\delta^+(S) := \{(i, j) \in E: i\in S, j\notin S\}$. If a connected component $C$ contains the depot $d_k$ the following procedure is used to find the largest violated sub-tour elimination constraint in $x^k(\delta^+(S))\geqslant y_i^k$. Given a strongly connected component $C$ that contains a depot $d_k$, $i\in C\setminus \{d_k\}$, and a fractional solution $(\bm{x}^{*},\bm{y}^{*})$, the most violated constraint of the form $x^k(\delta^+(S))\geq y_i^k$ can be obtained by computing a minimum $s-t$ cut on a capacitated directed graph $\bar{G}_k=(\bar{V}_k,\bar{E}_k)$, with $\bar{V}_k=V_k^{*}$. The vertex $s$ denotes the source vertex and $s=d_k$. The vertex $t$ denotes the sink vertex and $t=i$, and the edge set is given by $\bar{E}_k=E_k^{*}$. Every edge $(i,j) \in \bar{E}_k$ is assigned a capacity $x_{ij}^{k*}$. We now compute the minimum $s-t$ cut $(S,\bar{V}_k\setminus S)$ with $t\in\bar{V}_k\setminus S$. The vertex set $S'=\bar{V}_k\setminus S$ defines the most violated inequality if the capacity of the cut is strictly less than $y_i^{k*}$. Clearly, the targets $i$ with $y_i^{k*}=0$ need not be considered. This algorithm can be repeated for every vehicle to generate the violated sub-tour elimination constraints. Once the set $S'$ that defines a violated sub-tour elimination constraint is obtained, this constraint is added to the formulation and the branch-and-cut algorithm is continued.

\section{Computational Results} \label{sec:results}
We now present detailed computational results to corroborate the effectiveness of the two-stage stochastic approach to deal with uncertainty in service times, by comparing its solution with the deterministic EVP. All the simulation experiments were performed on a MacBook Pro with a $2.9$ GHz Intel Core $i5$ processor and $16$ GB RAM using CPLEX 12.7 as a mixed-integer linear programming solver. The branch-and-cut algorithm with the dynamic cut-generation  routine presented in Sec. \ref{sec:algorithm} was implemented in C++ using the callback functionality of CPLEX. The internal cut-generation routines of CPLEX were switched off and CPLEX was used only to manage the enumeration tree in the branch-and-cut algorithm. All computation times are reported in seconds and an upper limit of $3600$ seconds was imposed on each run of the algorithm. The performance of the algorithm was tested on instances generated from the traveling salesman problem library (TSPLIB) \cite{Reinelt1991}.

\subsection{Instance Generation} \label{subsec:generation}
We generated $13$ instances containing $29$ targets from one TSPLIB instance named \emph{bays29}. Since the objective of this paper is to present a two-stage framework to address stochasticity in path planning problems for heterogeneous UVs, we restrict the total number of instances to a small value and concentrate on its merits over the deterministic EVP. The number of vehicles, $n$, and the number of functional heterogeneous targets for every vehicle, $R_k$, in the single TSPLIB instance with $29$ targets is varied in the sets $\{1,2,3,4,5\}$ and $\{1,3,5\}$, respectively. For the instance with just one vehicle, no functional heterogeneous targets are present; this instance is merely used to illustrate the result in Prop. \ref{prop:onevehicle}. The depot locations for the vehicles and their desired heading angles at the vertices were uniformly randomly generated. The minimum turn radius of the vehicles were generated according to the following procedure: for each instance, the grid size $g$ was computed to be the maximum of the coordinates of all the vertices; now the minimum turn radius was computed using the formula $3\cdot k\cdot g/100$ where $k=1,\cdots,n$. The minimum length path for each pair of vertices was computed using the results of and Dubins \cite{Dubins1957}. We assign a name to each of the $13$ instances and the names conform to the format ``bays29-$n$-$f$'', where $n$ and $f$ are the number of vehicles and number of functional heterogeneous targets per vehicle, respectively. $100$ scenarios for the service times for each vehicle--target pair, $\tau_{ik}^\omega$, are generated randomly from a uniform distribution; each scenario is assumed to be equally likely and hence $p_{ik}^\omega = \frac{1}{|\Omega|}$, where $|\Omega| = 100$. The maximum allowable service time for each vehicle--target pair, $\bar{\tau}_{ik}$ is set to the average of $\bm{\tau}_{ij}^\omega$, $\omega \in \Omega$, offset by a random value from $[-3,3]$ and the penalties $\gamma_k$ is set to $1000$ for every $k$.

\subsection{Computation times} \label{subsec:times} 
The computation times for the algorithm to compute the optimal solution with $100$ scenarios for each of the $13$ instances is shown in Table \ref{tab:times}. For instances with around $30$ targets and $100$ scenarios, the branch-and-cut algorithm is observed to be reasonably fast and can compute the optimal solution within half a minute. We remark that no simulation experiments have been performed to test the scalability of the algorithm with increasing number of targets or number of scenarios. As elaborated in the Sec. \ref{subsec:related_work}, developing decomposition algorithms to solve the formulation in Sec. \ref{sec:formulation} for large instances of the problem with approximately $100$ targets and greater than $1000$ scenarios is a separate research topic; we delegate this algorithmic development for future work. 

\begin{table}[htbp]
\centering
\caption{Computation time in seconds with $100$ scenarios}
\label{tab:times}
\begin{tabular}{cc}
    \toprule
    instance name & {\centering computation time (sec)} \\
    \midrule 
    bays29-1-0 & 0.16 \\
    bays29-2-1 & 5.10 \\
    bays29-2-3 & 1.13 \\
    bays29-2-5 & 1.54 \\
    bays29-3-1 & 7.88 \\
    bays29-3-3 & 1.01 \\
    bays29-3-5 & 0.28 \\
    bays29-4-1 & 8.22 \\
    bays29-4-3 & 2.72 \\
    bays29-4-5 & 0.13 \\
    bays29-5-1 & 17.88 \\
    bays29-5-3 & 1.79 \\
    bays29-5-5 & 0.11 \\
    \bottomrule
\end{tabular}
\end{table}

\subsection{Branch-and-cut algorithm performance} \label{subsec:algo-performance}
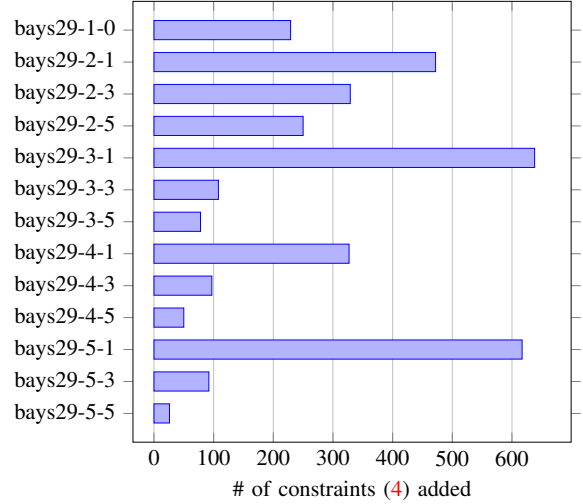
\begin{figure}[htbp]
\centering
\begin{tikzpicture}[scale=0.85]
\begin{axis}[
  xmajorgrids,
  xbar, 
  y=-0.5cm,
  bar width=0.3cm,
  enlarge y limits={abs=0.45cm},
  xlabel={\# of constraints \eqref{eq:sec} added},
  symbolic y coords={bays29-1-0,bays29-2-1,bays29-2-3,bays29-2-5,bays29-3-1,bays29-3-3,
  bays29-3-5,bays29-4-1,bays29-4-3,bays29-4-5,bays29-5-1,bays29-5-3,bays29-5-5},
  ytick=data,
  %nodes near coords, nodes near coords align={horizontal},
  ]
\addplot table[col sep=comma,header=false] {
229,bays29-1-0
472,bays29-2-1
329,bays29-2-3
250,bays29-2-5
638,bays29-3-1
108,bays29-3-3
78,bays29-3-5
327,bays29-4-1
97,bays29-4-3
50,bays29-4-5
617,bays29-5-1
92,bays29-5-3
26,bays29-5-5
};
\end{axis}
\end{tikzpicture}
\caption{Number of connectivity constraints \eqref{eq:sec} added by the branch-and-cut algorithm}
\label{fig:cuts}
\end{figure}

The main advantage of the branch-and-cut algorithm is that it can deal with the exponential number of sub-tour elimination or connectivity constraints (in Eq. \eqref{eq:sec}) by dynamically generating them as and when they are violated by the current feasible solution in the enumeration tree. This procedure of dynamically generating the constraints is found to be very effective since only a fraction of the total number of connectivity constraints are required for convergence to the optimal solution; this is corroborated by the results in Fig. \ref{fig:cuts}.

\subsection{Comparison to the EVP} \label{subsec:evp}
In this section, we explicitly compute the VSS, defined in Sec. \ref{subsec:vss}, for all the instances. The Table \ref{tab:vss} shows the cost of the solution obtained via the two-stage stochastic program ($\mathcal S^*$) and the cost obtained by using the solution of the EVP in the two-stage stochastic program ($\mathcal D^*$), respectively. The first entry in Table \ref{tab:vss} has a VSS value of zero, since it is a single vehicle variant of the problem (see Prop. \ref{prop:onevehicle}). The Fig. \ref{fig:vss-times} shows the extra computation time in seconds to solve the EVP when compared to the two stage stochastic formulation. This is not surprising because, the two-stage problem is a much more difficult problem as it accounts for uncertainty in the service times whereas the deterministic EVP accounts for the uncertainty only using the expectation. Nevertheless, efficient decomposition techniques can be utilized to reduce the gap within a stipulated runtime. 
\begin{table}[htbp]
\centering
\caption{Value of Stochastic Solution (VSS)}
\label{tab:vss}
\begin{tabular}{cccr}
    \toprule
    instance name & $\mathcal D^*$ & $\mathcal S^*$ & VSS \\
    \midrule  
    bays29-1-0	& 16,574.10	& 16,574.1	& 0.00 \\
    bays29-2-1	& 17,996.70	& 15,830.6	& 2,166.10 \\
    bays29-2-3	& 16,195.50	& 15,720.0	& 475.50 \\
    bays29-2-5	& 18,993.20	& 18,475.5	& 517.70 \\
    bays29-3-1	& 17,918.80	& 16,391.4	& 1,527.40 \\
    bays29-3-3	& 22,345.60	& 20,319.6	& 2,026.00 \\
    bays29-3-5	& 23,869.50	& 23,199.3	& 670.20 \\
    bays29-4-1	& 20,226.40	& 19,194.1	& 1,032.30 \\
    bays29-4-3	& 30,235.30	& 26,952.0	& 3,283.30 \\
    bays29-4-5	& 32,883.50	& 32,769.1	& 114.40 \\
    bays29-5-1	& 24,817.10	& 23,389.1	& 1,428.00 \\
    bays29-5-3	& 35,212.50	& 33,167.3	& 2,045.20 \\
    bays29-5-5	& 46,496.20	& 46,062.2	& 434.00 \\
    \bottomrule
\end{tabular}
\end{table}
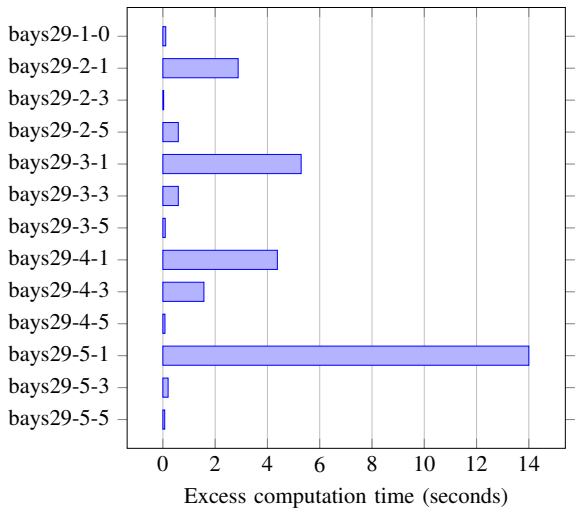
\begin{figure}[htbp]
\centering
\begin{tikzpicture}[scale=0.85]
\begin{axis}[
  xmajorgrids,
  xbar, 
  y=-0.5cm,
  bar width=0.3cm,
  enlarge y limits={abs=0.45cm},
  xlabel={Excess computation time (seconds)},
  symbolic y coords={bays29-1-0,bays29-2-1,bays29-2-3,bays29-2-5,bays29-3-1,bays29-3-3,
  bays29-3-5,bays29-4-1,bays29-4-3,bays29-4-5,bays29-5-1,bays29-5-3,bays29-5-5},
  ytick=data,
  %nodes near coords, nodes near coords align={horizontal},
  ]
\addplot table[col sep=comma,header=false] {
0.11,bays29-1-0
2.88,bays29-2-1
0.03,bays29-2-3
0.59,bays29-2-5
5.29,bays29-3-1
0.59,bays29-3-3
0.09,bays29-3-5
4.38,bays29-4-1
1.57,bays29-4-3
0.08,bays29-4-5
14.00,bays29-5-1
0.20,bays29-5-3
0.07,bays29-5-5
};
\end{axis}
\end{tikzpicture}
\caption{Additional computation time to solve the two-stage stochastic formulation when compared to the EVP}
\label{fig:vss-times}
\end{figure}

\begin{figure}
\centering
\includegraphics[scale=0.45]{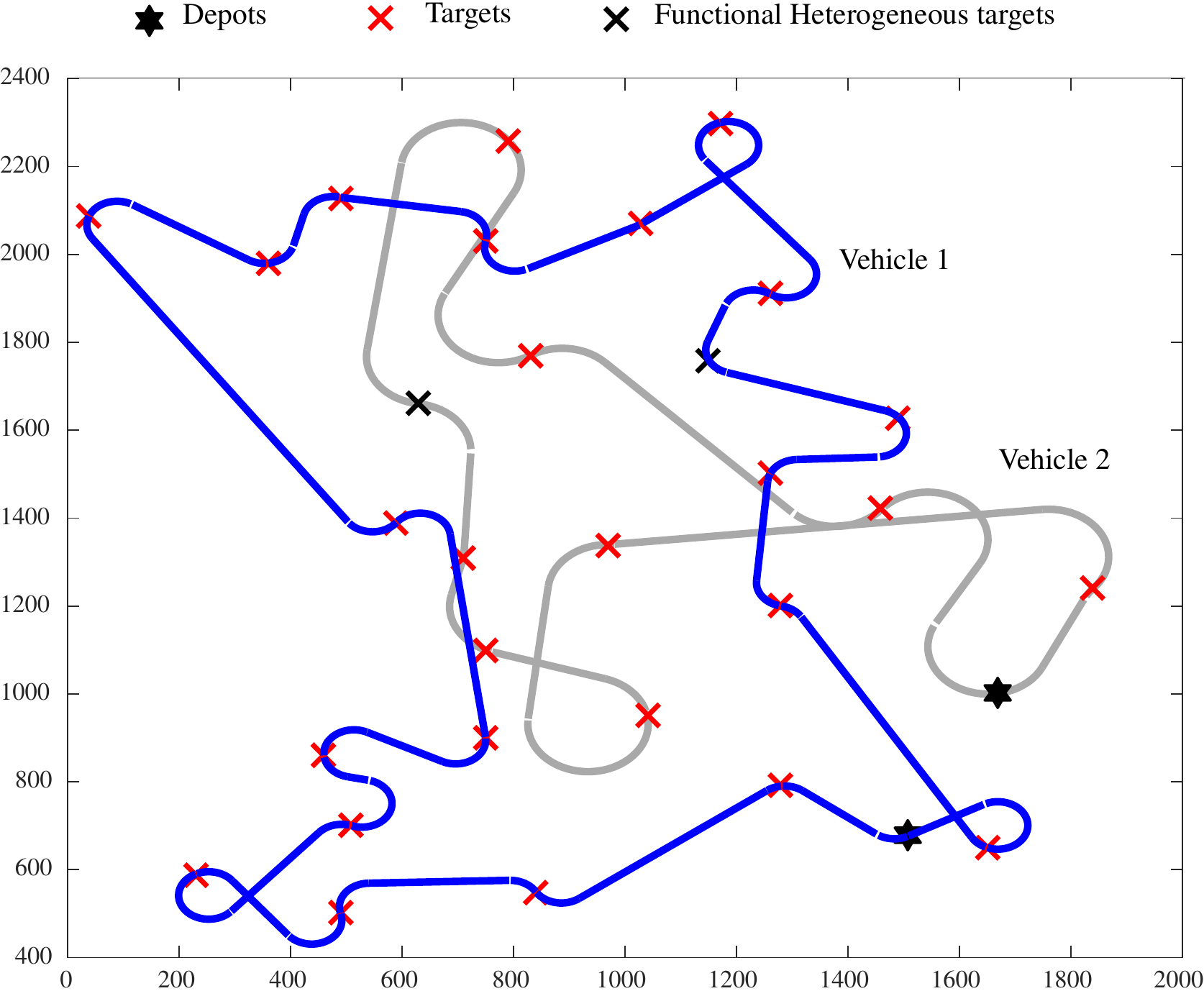}
\caption{Optimal paths computed by the two-stage stochastic formulation for the instance bays29-2-1}
\label{fig:plot-rs}
\end{figure}

\begin{figure}
\centering
\includegraphics[scale=0.45]{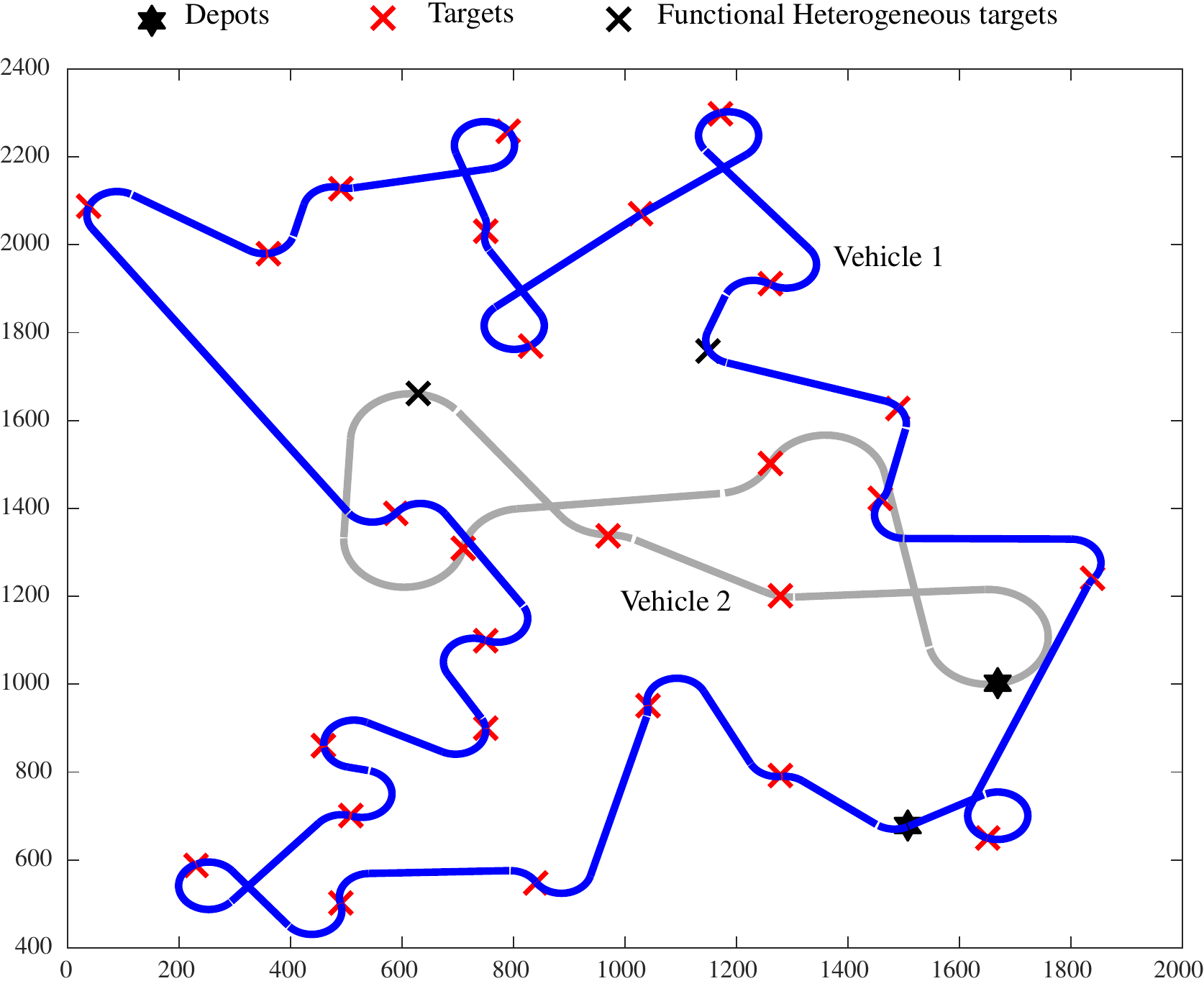}
\caption{Optimal paths computed by the EVP for the instance bays29-2-1}
\label{fig:plot-evp}
\end{figure}

Fig. \ref{fig:plot-rs} and \ref{fig:plot-evp} show the optimal paths obtained from the two-stage stochastic program and the EVP for the instance bays29-2-1, respectively; the instance has two vehicles and one functional heterogeneous target per vehicle. The traveling cost for the vehicles (the first-stage costs) are $14,975.4$ and $14,004.7$ for the two-stage stochastic program and the EVP for the instance bays29-2-1, respectively. Though the EVP produces paths for the vehicles whose total cost is less than the two-stage stochastic program, the expected penalty incurred by the EVP solution would be much higher than that produced the two-stage stochastic program (see row $2$ in Table \ref{tab:vss}).

\section{Conclusion} \label{sec:conclusion}
This paper formulates the multiple depot heterogeneous multiple vehicle problem with random service times for each vehicle--target pair as a two-stage stochastic problem with recourse. The formulation is compared with the deterministic expected value problem formulation and theoretical conditions under which both the formulations are equivalent are derived. The two formulations are compared using the value of the stochastic solution (VSS), which measures the cost of including uncertainty in the decision making process. A branch-and-cut algorithm to solve the two-stage stochastic formulation to optimality is then developed followed by extensive computational results. Future work can include developing computationally efficient algorithms using decomposition--based techniques and scale the algorithm both in terms of the number of targets and in terms of the number of scenarios. 

\bibliographystyle{IEEEtran}
\bibliography{references}

\end{document}